\documentclass[journal,twoside,web]{IEEEtran}
\usepackage{cite}
\usepackage{amsmath,amssymb,amsfonts}
\usepackage{algorithmic}
\usepackage{graphicx}
\usepackage{textcomp}
\usepackage[dvipsnames]{xcolor}
\usepackage{hyperref}
\usepackage{amsmath}
\usepackage{cleveref}
\usepackage{xcolor}
\usepackage{epsfig}
\usepackage{graphicx}

\usepackage{amssymb}
\usepackage{amsmath}

\usepackage{amsthm}
\usepackage{mathrsfs}

\usepackage{adjustbox} 

\def\R{\mathbb{R}}

\def\G{\mathcal{G}}
\def\S{\mathbb{S}}

\def\T{\mathcal{T}}
\def\bx{\mathbf{x}}
\newcommand{\Sk}{\widehat{\mathbb{S}}^{(k)}}
\newcommand{\Vk}{\widehat{\mathbb{V}}^{(k)}}

\newtheorem{theorem}{Theorem}[section]
\newtheorem{proposition}[theorem]{Proposition}

\begin{document}
\title{Linear optimal transport subspaces for point set classification}
\author{Mohammad Shifat-E-Rabbi, Naqib Sad Pathan, Shiying Li, Yan Zhuang, Abu Hasnat Mohammad Rubaiyat, and Gustavo K Rohde
\thanks{M.S.E. Rabbi is with the Department of Electrical and Computer Engineering, North South University, Dhaka, Bangladesh (e-mail: rabbi.mohammad@northsouth.edu).}
\thanks{N.S. Pathan is with the Imaging and Data Science Laboratory and the Department of Electrical and Computer Engineering, University of Virginia, Charlottesville, VA, USA (e-mail: qpb3vt@virginia.edu).}
\thanks{S. Li is with the Department of Mathematics, University of North Carolina - Chapel Hill, NC, USA (e-mail: shiyl@unc.edu).}
\thanks{Y. Zhuang is with the Department of Radiology and Imaging Sciences, National Institutes of Health Clinical Center, MD, USA (e-mail: yan.zhuang2@nih.gov).}
\thanks{A.H.M. Rubaiyat is with the Imaging and Data Science Laboratory and the Department of Electrical and Computer Engineering, University of Virginia, Charlottesville, VA, USA (e-mail: ar3fx@virginia.edu).}
\thanks{G.K. Rohde is with the Imaging and Data Science Laboratory, the Department of Biomedical Engineering, and the Department of Electrical and Computer Engineering, University of Virginia, Charlottesville, VA, USA (e-mail: gustavo@virginia.edu).}
\thanks{M.S.E. Rabbi (rabbi.mohammad@northsouth.edu) is the corresponding author.}}

\maketitle

\begin{abstract}

Learning from point sets is an essential component in many computer vision and machine learning applications. Native, unordered, and permutation invariant set structure space is challenging to model, particularly for point set classification under spatial deformations. Here we propose a framework for classifying point sets experiencing certain types of spatial deformations, with a particular emphasis on datasets featuring affine deformations. Our approach employs the  Linear Optimal Transport (LOT) transform to obtain a linear embedding of set-structured data. Utilizing the mathematical properties of the LOT transform, we demonstrate its capacity to accommodate variations in point sets by constructing a convex data space, effectively simplifying point set classification problems. Our method, which employs a nearest-subspace algorithm in the LOT space, demonstrates label efficiency, non-iterative behavior, and requires no hyper-parameter tuning. It achieves competitive accuracies compared to state-of-the-art methods across various point set classification tasks. Furthermore, our approach exhibits robustness in out-of-distribution scenarios where training and test distributions vary in terms of deformation magnitudes. 


\end{abstract}



\section{Introduction}
Point sets provide valuable insights about object geometry, making them useful for a variety of applications, including object detection, recognition, segmentation, and tracking in fields such as robotics, autonomous vehicles, virtual reality, and computer vision, among others \cite{qi2017pointnet,zhao2019pointweb,li2018pointcnn,chen2017multi,zhou2018voxelnet}. They represent the surface geometry of an object in an N-dimensional space as a set of points, obtained using various scanning technologies such as LiDAR or photogrammetry \cite{xu2021toward,wang2020computational}, or by sampling a continuous probability density function over an N-D space \cite{zhou20213d}. However, modeling the set structure space for classification presents significant challenges due to the sparsity and noise in data, the accumulation of spatial deformations (such as affine deformations) in real-world point set data, and the high dimensionality of point sets, among other factors \cite{pomerleau2015review,wang2020cascaded,qi2017pointnet,zeng20193d}. Furthermore, defining a metric or distance function for point set classification is challenging due to the permutation invariant nature of point sets resulting from the arbitrary order of points in a set \cite{qi2017pointnet,lu2021slosh}. However, despite these challenges, there has been a growing interest in developing new algorithms and techniques for point set classification. 

In recent years, several research efforts have focused on point set classification, resulting in the development of various methods to address challenges in this area. Over the last few decades, point set classification methods have evolved from relying on feature engineering \cite{radenovic2018fine,wang2020deep,acharya2018covariance,zhang2019fspool} to utilizing deep neural networks to learn representations and use them in classification tasks \cite{li2018pointcnn,qi2017pointnet,wang2019dynamic}. Neural networks have emerged as a leading classification framework for point sets, providing end-to-end learning capabilities and eliminating the need for hand-crafted feature engineering. They have demonstrated to achieve high accuracy in several classification tasks, and are also suitable for parallel implementation using graphical processing units (GPUs) \cite{li2018pointcnn,qi2017pointnet,wang2019dynamic}. However, the effectiveness of neural network-based methods is often limited by their high data requirements \cite{liu2019deep}, high computational costs \cite{qi2017pointnet}, and vulnerability to out-of-distribution samples, e.g., adversarial attacks \cite{liao2018defense,basu2014detecting,shifat2021radon}. %

While the conventional approach to modeling point sets involves direct processing of their coordinates, an alternative and less commonly used method is to represent a point set as a deformation of another point set \cite{kolouri2017optimal}. To address this challenge, point set deformation models have been developed utilizing the mathematics of optimal mass transport \cite{kolouri2017optimal,wang2013linear}. These models treat a point set as a smooth, nonlinear, and invertible transformation of a reference point set structure. The estimation of such models can be facilitated through the use of the linear optimal transport (LOT) transform, which has found applications in various fields \cite{wang2013linear}. The LOT of a point set provides as a linear embedding for that point set which can be used to compare with other point set data \cite{wang2013linear}. The LOT transform has been combined with various machine learning techniques and has been used in many applications \cite{wang2013linear,basu2014detecting}.

This paper introduces a new method for classifying point sets by expanding upon the LOT-based modeling frameworks. We start by introducing a transport generative model to define point set classes, where class elements can be conceived as instances of an unknown template point set pattern under the effect of unknown spatial deformations. Using the mathematical properties of the LOT transform, we establish that these point set classes, under our generative model (with certain conditions on spatial deformations), can be constructed as convex subspaces in the LOT space, which are capable of accommodating the variations in point set data. Subsequently, we propose a nearest subspace-based classifier in the LOT space for classifying point sets under the given generative model. Our model is also capable of mathematically encoding invariances by integrating mathematical knowledge of deformations known to be present in the data.  In our experiments, we particularly focus on datasets experiencing affine deformations and demonstrate the effectiveness of our method compared to several state-of-the-art methods. Our approach exhibits particular strength in situations characterized by limited training data and in the challenging out-of-distribution setting, where the training and test distributions differ in terms of deformation magnitudes.


\section{Preliminaries}
\label{preliminaries_sec_aim2}


\subsection{Linear optimal transport embeddings}
The fundamental principle of optimal transport theory relies on quantifying the amount of effort (measured as the product of mass and distance) required to rearrange one distribution to another, which gives rise to the Wasserstein metric between distributions. In the present study, we utilize a linearized version of this metric, as outlined in \cite{wang2013linear}, which is constructed formally through a tangent space approximation of the underlying manifold.

Following the construction in \cite{wang2013linear}, we define the linear optimal transport transform for probability measures in $\mathcal{P}_2(\R^L)$, which is the set of absolutely continuous measures with bounded finite second moments and densities \footnote{Any $\mu\in \mathcal{P}_2(\R^L)$ has the following two properties (i) bounded second moment, i.e. $\int\|x\|^2 d\mu(x)<\infty$; (ii) absolute continuity with respect to the Lebesgue measure on $\R^L$ with bounded density, i.e., $\mu$ has a density function $f_{\mu}$ defined on $\R^L$ with $\|f_{\mu}\|_{\infty}<\infty$. 
 }. For simplicity, let us fix a reference measure $\sigma$ as the Lebesgue measure on a convex compact set of $\R^L$.
 Thanks to Brenier's theorem \cite{brenier1991}, there is a unique minimizer $T_{\sigma}^\mu$ to the following optimal transportation problem
\begin{equation}
  \min \limits_{T_{\sharp}\sigma = \mu}\int_{\R^L} \|x-T(x)\|^2d\sigma(x),
\end{equation}
where the push-forward (transport) relation $T_{\sharp}\sigma = \mu$ is defined via $\mu(B) = \sigma(T^{-1}(B))$ for any measurable set $B\subseteq \R^L$.
The linear optimal transport (LOT) transform is given by the following correspondence 
\begin{equation}
    \mu \mapsto T_{\sigma}^{\mu},
\end{equation}
where each probability measure $\mu$ is identified with the optimal transport map $T_{\sigma}^{\mu}:\R^L\rightarrow \R^L$ from a fixed reference $\sigma$ to $\mu$, which lies in a linear space. This square-root of the minimum is called the Wasserstein-2 distance between $\sigma$ and $\mu$ \cite{villani2003topics}. The LOT metric between two probability distributions $\mu,\nu\in \mathcal{P}_2(\R^L)$ is \footnote{Note that $\|T\|_{\sigma}:= \Big(\int_{\R^L}\|T(x)\|^2 d\sigma(x)\Big)^{1/2}$.}
\begin{equation}
    d_{\textrm{LOT}}(\mu,\nu):= \|T_{\sigma}^{\mu}-T_{\sigma}^{\nu}\|_{\sigma}. 
\end{equation} 
For simplicity, we denote $\widehat \mu$ as the LOT transform of $\mu$, i.e.,  $\widehat \mu= T_{\sigma}^\mu$ where $\sigma$ is fixed.

It turned out the linearization ability of LOT is closely related to the scope of the following so-called composition property \cite{aldroubi20,moosmuller2023linear}
\begin{equation}\label{CompositionProp}
    T_{\sigma}^{g_{\sharp}\mu} = g\circ T_{\sigma}^{\mu},
\end{equation}
where $g\in \T_L$, and $\T_L$ is the set of all diffeomorphisms from $\R^L$ to $\R^L$. In particular, given a convex $\mathcal{G}\subseteq \T_L$, the LOT embedding of deformed measures via maps in $\mathcal{G}$ become convex \footnote{ Note in general $\mathcal{G}_{\sharp}\mu$ is not convex as $(\lambda_1g_1+\lambda_2g_2)_{\sharp}\mu \neq \lambda_1{g_1}_{\sharp}\mu + \lambda_2{g_2}_{\sharp}\mu$.} if all $g\in \mathcal{G}$ satisfies the above composition property \eqref{CompositionProp}, which is shown more formally below.
\begin{proposition}[Lemma A.2 in \cite{moosmuller2023linear}] \label{PropConvex}
Let $\mathcal{G}\subseteq \T_L$ be convex. Given $\mu\in \mathcal{P}_2(\R^L)$, define $\mathcal{G}_{\sharp}\mu := \{g_{\sharp}\mu: g\in \mathcal{G}\}$. If $\forall g\in \mathcal{G}$, \eqref{CompositionProp} holds, 
then $\widehat {\mathcal{G}_{\sharp}\mu}:= \{\widehat \nu: \nu \in \mathcal{G}_{\sharp}\mu\}$ is convex in the LOT transform domain.
\end{proposition}

When the dimension $L\geq 2$, it is shown in \cite{aldroubi20} that  $g$ can only be ``basic" transformations (more specifically,  translations or isotropic scalings or their compositions) for the composition property \eqref{CompositionProp} to hold for arbitrary $\mu$'s. Luckily, \cite{moosmuller2023linear} proposes an approximate composition property for perturbations of the aforementioned basic transformations, the set of which we denote as $\mathcal{A} =\{h(x)=ax+b: a>0, b\in\R^L\}$. 

\vspace{1em}
\noindent{\bf{Property 1} (Approximate composition,  p.388 in \cite{moosmuller2023linear}\footnote{This property is referred as $\delta$-compatibility in \cite{moosmuller2023linear}.})} Let $\epsilon\geq 0$ and $\mu\in \mathcal{P}_2(\R^L)$. Let $g \in \T_L$ such that $\|g-h\|\leq \epsilon$ for some $h\in \mathcal{A}$. Then there exists some $\delta$ such that 
\begin{equation}
    \|T_{\sigma}^{g_{\sharp}\mu}- g\circ T_{\sigma}^{\mu} \|_{\sigma} < \delta,
\end{equation}

\noindent{Remark:} Using the $\widehat{\mu}$ notation for LOT transform of $\mu$, we have 
\begin{equation}
\label{eqn_appconvex}
\|\widehat{g_{\sharp}\mu}- g\circ \widehat \mu \|_{\sigma} < \delta.
\end{equation}

With the above approximate composition property,  one can show the following approximate convexity analog of Proposition \ref{PropConvex} using Lemma A.3, A.4 of \cite{moosmuller2023linear}:
\begin{proposition}\label{PropApproxConvex_aim2}
    Let $\epsilon\geq 0$ and  $\mathcal{G}\subseteq \T_L$ be convex such that for any $g\in \mathcal{G}$, there exists some $h\in \mathcal{A}$ such that $\|g-h\|\leq \epsilon$. Given $\mu\in \mathcal{P}_2(\R^L)$, we have $\widehat {\mathcal{G}_{\sharp}\mu}:= \{\widehat \nu: \nu \in \mathcal{G}_{\sharp}\mu\}$ is $2\delta$-convex in the LOT transform domain, where $\delta$ is given in the above approximate composition property.  In particular, for any $c\in [0,1]$ and $\widehat{{g_1}_{\sharp}\mu}, \widehat{{g_2}_{\sharp}\mu}\in \widehat {\mathcal{G}_{\sharp}\mu}$ ($g_1, g_2\in \mathcal{G}$), 
    \begin{equation}
        \|(1-c)\widehat{{g_1}_{\sharp}\mu} + c\widehat{{g_2}_{\sharp}\mu} - \widehat{{g_c}_{\sharp}\mu}\|< 2\delta,
    \end{equation}
    where $g_c = (1-c)g_1+cg_2\in \mathcal{G}$.
\end{proposition}

\subsection{Discrete implementation for point sets}
For the analysis of discrete point set data, a discrete version of the Linear Optimal Transport (LOT) embedding is required. In this particular case, both the reference $\sigma$ and target $\mu$ are chosen as discrete probability measures, represented by point sets in $\R^L$. A point set in a $L$-dimensional space is a finite set of points in $\R^L$. A point set $\Omega_s$ with $N$ points can be thought as the image of an injective map $s: \{1,\cdots ,N\}\rightarrow \R^L$ \footnote{Note that a point set may be associated with many injective maps, e.g. the image sets of $s\circ \gamma$ and $s$ are the same for any permutation $\gamma$.}.  Given  a point set $\Omega_s$  with $N$ points, we define a discrete probability distribution associated with the point set as
\begin{align}
    P_s:=\frac{1}{N}\sum_{\bx\in \Omega_s}\delta_{\bx} = \frac{1}{N}\sum_{i=1}^N\delta_{s(i)}.
\end{align}
Given a diffeomorphism $g\in \T_L$,  the push-forward distribution of $P_s$ under $g$ is given as 
\begin{align}
    g_{\#}P_s:=\frac{1}{N}\sum_{\bx\in \Omega_s}\delta_{g(\bx)}=\frac{1}{N}\sum_{i=1}^N\delta_{g(s(i))} = P_{g\circ s}.
\end{align}

Let $\mathcal{F_{N,L}}$ denote the collection of injective maps from $\{1,\cdots, N\}$ to $\R^L$.  Given $s, r\in \mathcal{F_{N,L}}$, the optimal transportation (Wasserstein-2) distance  between associated distributions $P_s$ and $P_r$ can be obtained by solving the linear programming problem given below:
\begin{align}\label{linearProgram}
d_W^2(P_s,P_r)=\min_{\pi\in \R^{N\times N}}\sum_{i=1}^N\sum_{j=1}^N\pi_{ij}|s(i)-r(j)|^2
\end{align}
where $ \pi_{ij}\geq0$, and $\sum_{i=1}^N\pi_{ij}=\sum_{j=1}^N\pi_{ij}=1/N$ for all $i, j = 1,\cdots, N$. Let us fix some $r \in \mathcal{F_{N,L}}$ and use $P_r$ as a reference.
It turned out  that any minimizer matrix $\pi^{*}$ to the optimal transport problem in \eqref{linearProgram} is a permutation matrix\cite{villani2003topics}. In other words, there is a permutation $\sigma_s^{*}: \{1,\cdots, N\}\rightarrow \{1,\cdots, N\}$ such that 
\[\pi^{*}_{ij}= 
\begin{cases}
1/N\quad & \textrm{if}~  j = \sigma_s^*(i)\\
0 \quad & \textrm{otherwise}
\end{cases}. 
\]
Hence with $r$ being fixed, an optimal transport map between $P_r$ and $P_s$ can be  determined by $\sigma^*_s$ and $s$. 
The LOT transform for $P_s$  is defined as \cite{wang2013linear} \footnote{Note one can write $s\circ \sigma^*_s = \begin{bmatrix}
s(\sigma_s^*(1)),\cdots,
s(\sigma_s^*(N))
\end{bmatrix}^T.$ Note also that $\sigma^*_s$ may not be unique in general, we follow the implementation in \cite{wang2013linear} to estimate one of them.}
\begin{equation}
    \widehat{P}_{s}:  = s\circ \sigma^*_s, \label{eq_lot_aim2}
\end{equation} 
and the LOT distance between two point set measures is 
\begin{equation}
    d_{\textrm{LOT}}(P_s, P_q) := ||\widehat P_s -\widehat P_q||,
\end{equation}
where $s, q \in \mathcal{F_{N,L}}$.


\section{Transport based Classification Problem statement}
In this section, we present a generative model-based problem statement for the type of classification problems we discuss in this paper, building upon the preliminaries established earlier. Our focus is on point set classification, where every class can be viewed as a collection of instances of a prototype point set pattern (a template) observed under unknown spatial deformations. To formalize this concept, we introduce a generative model that provides a formal approach to characterizing point set data of this type.

\vspace{1em}
\noindent{\bf{Generative model}}
Let $\G_L\subset\T_L$ be a set of smooth one-to-one transformations in an $L$-dimensional space. The mass-preserving generative model for the $k$-th class is defined to be the set
\begin{align}
\label{gen_model_aim2}
    \S^{(k)}=\left\{P_{s_j^{(k)}}|P_{s_j^{(k)}}=g_{j\#}P_{\varphi^{(k)}},~\forall g_j\in\G_L\right\}
\end{align}
where $P_{\varphi^{(k)}}$ corresponds to the point set distribution of the prototype template pattern for the $k$-th class and $P_{s_j^{(k)}}$ represents the point set distribution of the $j$-th sample from the $k$-th class in $\S^{(k)}$. With these definitions, we can now construct a formal mathematical description for the generative model-based problem statement for point set classification. 


\vspace{1em}
\noindent{\bf{Classification problem:}} Let the set of point set distributions $\S^{(k)}$ are given as in equation~\eqref{gen_model_aim2}. Given training samples $\{P_{s_1^{(1)}},P_{s_2^{(1)}},\cdots\}$ (class 1), $\{P_{s_1^{(2)}},P_{s_2^{(2)}},\cdots\}$ (class 2), $\cdots$ as training data, determine the class of an unknown distribution $P_s$. 

Note that the generative model in equation~\eqref{gen_model_aim2} describes set-structured point set data, which makes it challenging to compare point sets due to their permutation-invariant nature. The generative model above is also not guaranteed to be convex, presenting challenges for effective classification using machine learning techniques. In the subsequent sections, we present solutions to the above classification problem at first by restructuring the point clouds by providing linear optimal transport (LOT) embeddings for them and then by approximating the resulting convex spaces with subspaces as done in many image \cite{shifat2021radon,shifat2023invariance}, signal \cite{rubaiyat2022nearest,rubaiyat2022end}, and gradient distribution \cite{zhuang2022local} classification problems.

\section{Proposed solution}
The LOT transform, which was previously described in section~\ref{preliminaries_sec_aim2}, can significantly simplify the classification problem described earlier by providing a convex linear embedding for the set-structured point set data. Let us first investigate the generative model in equation~\eqref{gen_model_aim2} in the LOT transform space. Applying the approximate composition property (equation~\eqref{eqn_appconvex}) to the generative model in equation~\eqref{gen_model_aim2}, we have the LOT-space generative model as follows:
\begin{align}
\label{gen_model_aim2_trans}
    \Sk=\left\{\widehat{P}_{s_j^{(k)}}|\widehat{P}_{s_j^{(k)}}=g_{j}\circ\widehat{P}_{\varphi^{(k)}},~\forall g_j\in\G_L\right\}
\end{align}

In this context, $\widehat{P}{s_j^{(k)}}$ and $\widehat{P}{\varphi^{(k)}}$ refer to the LOT embeddings of $P_{s_j^{(k)}}$ and $P_{\varphi^{(k)}}$, respectively, with respect to a reference structure $P_r$ (see equation~\eqref{eq_lot_aim2}). Based on the preliminary results presented in Section~\ref{preliminaries_sec_aim2} (Property~1, Proposition\ref{PropApproxConvex_aim2}, and other results), it is possible to establish the convexity of the set $\Sk$ up to a certain bound, subject to certain constraints. Furthermore, we can show that when $\S^{(k)}\cap\S^{(p)}=\varnothing$, the intersection of $\Sk$ with $\widehat{\S}^{(p)}$ is empty \cite{shifat2021radon}. 

\subsection{Training phase}
Based on the aforementioned theoretical discussions, we put forward a straightforward non-iterative training approach for the classification method. This involves computing a projection matrix that maps each sample in the LOT space onto the subspace $\Vk$ (as outlined in \cite{shifat2021radon}), generated by the 2$\delta$-convex set $\Sk$. Specifically, we estimate the projection matrix by applying the following procedure:
\begin{align}
    \Vk=\mbox{span}\left(\Sk\right)=\{\sum_{j\in J}\alpha_j\widehat{P}_{s_j^{(k)}}|\alpha_j\in\R,~J\mbox{ is finite}\}.\nonumber
\end{align}

Given a set of sample training data, denoted as $\{P_{s_1^{(k)}},P_{s_2^{(k)}},\cdots\}$, the first step in our proposed method is to apply the LOT transform on them using a reference distribution $P_{r^{(k)}}$. This results in the generation of transformed samples, denoted as $\{\widehat{P}_{s_1^{(k)}},\widehat{P}_{s_2^{(k)}},\cdots\}$. The reference distribution $P_{r^{(k)}}$ is obtained by selecting a point set at random from the training set, followed by the introduction of random perturbations. Subsequently, we estimate $\Vk$ as follows:
\begin{align}
  \Vk= \mbox{span} \{\widehat{P}_{s_1^{(k)}},\widehat{P}_{s_2^{(k)}},\cdots\}.
\end{align}

The proposed method also provides a structure to mathematically encode invariances with respect to deformations that are known to be present in the data \cite{shifat2021radon,shifat2023invariance}. In this paper, we prescribe methods to encode invariances with respect to a set of affine transformations: translation, isotropic and anisotropic scaling, and shear. Detailed descriptions of the deformation types used for encoding invariances and the corresponding methodologies are explained as follows:
\begin{enumerate}
    \item Translation: Let $g(\mathbf{x})= \mathbf{x}+\mathbf{x_0}$ be the translation by $\mathbf{x}_0=((\mathbf{x}_0)_1,(\mathbf{x}_0)_2,\cdots,(\mathbf{x}_0)_L)\in \R^L~\mbox{and}~P_{s_g}=g_\#P_s$. Using equation~\eqref{eqn_appconvex}, we have that $\widehat{P}_{s_g}=\widehat{g_\#P_s}\approx g\circ\widehat{P}_s=\widehat{P}_s+\mathbf{x}_0,~\mbox{where}~\widehat{P}_s=((\widehat{P}_s)_1,(\widehat{P}_s)_2,\cdots,(\widehat{P}_s)_L)$. Consequently,
    \begin{align}
        &\widehat{P}_{s_g}\approx\widehat{P}_s+\mathbf{x}_0=\widehat{P}_s+\left((\mathbf{x}_0)_1,(\mathbf{x}_0)_2,\cdots,(\mathbf{x}_0)_L\right)\nonumber\\
        &=\widehat{P}_s+(\mathbf{x}_0)_1\left(1,0,0,\cdots\right)+(\mathbf{x}_0)_2\left(0,1,0,\cdots\right)+\cdots\nonumber\\
        &+(\mathbf{x}_0)_L\left(0,0,\cdots,1\right).\nonumber
    \end{align} 
    Therefore, as in \cite{shifat2021radon,shifat2023invariance}, we define the spanning set for translation as 
    \begin{align}
        &\mathbb{U}_T=\{u_{t}(1),u_{t}(2),\cdots,u_{t}(L)\},~\mbox{where}\nonumber\\
        &u_{t}(1)=(1,0,0,\cdots), u_{t}(2)=(0,1,0,\cdots), \cdots,\nonumber\\
        &u_{t}(L)=(0,0,\cdots,1).\nonumber
    \end{align}
    
    \item Isotropic scaling: Let $g(\mathbf{x})=a\mathbf{x}$ be the normalized isotropic scaling of $P_s$ by $a$, where $a\in\R_{+}$ and $P_{s_g}=g_\#P_s$. Using equation~\eqref{eqn_appconvex}, we have that $\widehat{P}_{s_g}\approx g\circ\widehat{P}_s=a\widehat{P}_s$. As in \cite{shifat2021radon,shifat2023invariance}, an additional spanning set for isotropic scaling is not required as the subspace containing $\widehat{P}_s$ naturally contains its scalar multiplication $a\widehat{P}_s$. Therefore, the spanning set for isotropic is defined as $\mathbb{U}_{D_0}=\varnothing$.
    
    \item Anisotropic scaling: Let $g(\mathbf{x})=\breve{\mathcal{D}}\mathbf{x}$ be the normalized anisotropic scaling of $P_s$, where $\breve{\mathcal{D}}=\begin{bmatrix}a_1,&0,&\cdots\\0,&a_2,&\cdots\\ \vdots &\vdots &\ddots\end{bmatrix}$, $a_i\neq a_j$, $a_i\in\R_{+}$, and $P_{s_g}=g_\#P_s$. Using equation~\eqref{eqn_appconvex}, we have that $\widehat{P}_{s_g}\approx g\circ\widehat{P}_s=\breve{\mathcal{D}}\widehat{P}_s=\left(a_1(\widehat{P}_s)_1,a_2(\widehat{P}_s)_2,\cdots,a_L(\widehat{P}_s)_L\right)$. Consequently,
    \begin{align}
        &\widehat{P}_{s_g}\approx\breve{\mathcal{D}}\widehat{P}_s=a_1((\widehat{P}_s)_1,0,0,\cdots)+a_2(0,(\widehat{P}_s)_2,0,\cdots)+\nonumber\\
        &\cdots+a_L(0,0,\cdots,(\widehat{P}_s)_L).\nonumber
    \end{align}
    Therefore, the spanning set for anisotropic scaling is defined as 
    \begin{align}
    &\mathbb{U}_D=\{u_{d}(1),u_{d}(2),\cdots,u_{d}(L)\},~\mbox{where}\nonumber\\
    &u_{d}(1)=((\widehat{P}_s)_1,0,0,\cdots), u_{d}(2)=(0,(\widehat{P}_s)_2,0, \cdots),\nonumber\\
    &\cdots, u_{d}(L)=(0,0,\cdots,(\widehat{P}_s)_L).\nonumber
    \end{align}
    
    \item Shear: Let $g(\mathbf{x})=\mathcal{H}\mathbf{x}$ be the normalized shear of $P_s$, where $\mathcal{H}=\begin{bmatrix}1,&k_{12},&\cdots\\k_{21},&1,&\cdots\\ \vdots &\vdots &\ddots\end{bmatrix}$ and $P_{s_g}=g_\#P_s$. Here, the shear matrix $\mathcal{H}$ is constructed using the shear factors $k_{ij}\in\mathbb{R}$, which are located at the non-diagonal positions of $\mathcal{H}$. Using equation~\eqref{eqn_appconvex}, we have that $\widehat{P}_{s_g}\approx g\circ\widehat{P}_s=\mathcal{H}\widehat{P}_s=((\widehat{P}_s)_1+k_{12}(\widehat{P}_s)_2+k_{13}(\widehat{P}_s)_3+\cdots,(\widehat{P}_s)_2+k_{21}(\widehat{P}_s)_1+k_{23}(\widehat{P}_s)_3+\cdots,\cdots,(\widehat{P}_s)_L+k_{L1}(\widehat{P}_s)_1+k_{L2}(\widehat{P}_s)_2+\cdots)$. Consequently,
\begin{align}
    &\widehat{P}_{s_g}\approx\mathcal{H}\widehat{P}_s=\widehat{P}_s+k_{12}((\widehat{P}_s)_2,0,0,\cdots)+\nonumber\\
    &k_{13}((\widehat{P}_s)_3,0,0,\cdots)+k_{14}((\widehat{P}_s)_4,0,0,\cdots)+\cdots+\nonumber\\
    &k_{21}(0,(\widehat{P}_s)_1,0,\cdots)+k_{23}(0,(\widehat{P}_s)_3,0,\cdots)+\cdots+\nonumber\\
    &k_{L1}(0,0,\cdots,(\widehat{P}_s)_1)+k_{L2}(0,0,\cdots,(\widehat{P}_s)_2)+\cdots\nonumber. 
\end{align}
Therefore, the spanning set for shear is defined as 
\begin{align}
&\mathbb{U}_S=\{u_{s}(1,2),u_{s}(1,3),\cdots,u_{s}(L,L-1)\},~\mbox{where}\nonumber\\
&u_{s}(1,2)=((\widehat{P}_s)_2,0,0,\cdots), u_{s}(1,3)=((\widehat{P}_s)_3,0,0,\cdots),\nonumber\\
&\cdots,u_{s}(L,L-1)=(0,0,\cdots,(\widehat{P}_s)_{L-1}).\nonumber
\end{align}

\end{enumerate}

Finally, in light of the preceding discussion, we can approximate the enriched subspace $\Vk_E$ as
\begin{align}
  \Vk_E= \mbox{span} \left(\{\widehat{P}_{s_1^{(k)}},\widehat{P}_{s_2^{(k)}},\cdots\}\cup \mathbb{U}_A\right),
\end{align}
where $\mathbb{U}_A=\mathbb{U}_T\cup\mathbb{U}_{D_0}\cup\mathbb{U}_D\cup\mathbb{U}_S$.

\begin{figure}[]
	\centering
	\includegraphics[width=0.45\textwidth]{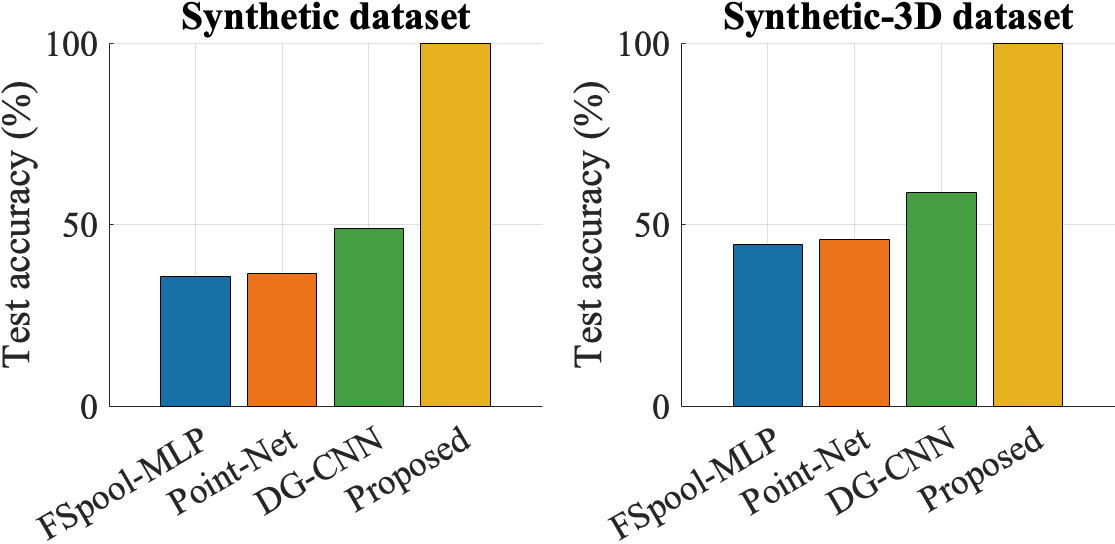}
	\caption{Percentage test accuracy comparison of different methods on synthetic datasets.}
	\label{resfig00_aim2}
\end{figure}

\subsection{Testing phase} 
To classify a given test sample $P_s$, we first apply the LOT transform to $P_s$ to obtain its corresponding LOT space representation $\widehat{P}_{s,r^{(k)}}$ with respect to the reference $P_{r^{(k)}}$ (which was pre-selected duing the training phase). 
Assuming that the test samples originate from the generative model presented in equation~\eqref{gen_model_aim2} (or equation~\eqref{gen_model_aim2_trans}), we can determine the class of an unknown test sample $P_s$ using the following expression: 
\begin{align}
\arg\min_kd^2\left(\widehat{P}_{s,r^{(k)}},\Vk_E\right)
\end{align}
where $d(\cdot,\cdot)$ is the distance between the test sample and the trained subspaces in the LOT transform space. We can estimate the distance between $\widehat{P}_{s,r^{(k)}}$ and the trained subspaces using $d^2\left(\widehat{P}_{s,r^{(k)}},\Vk_E\right)\sim ||\widehat{P}_{s,r^{(k)}}-B^{(k)}B^{(k)T}\widehat{P}_{s,r^{(k)}}||^2_{L_2}$, where the matrix $B^{(k)}$ contains the basis vectors of the subspace $\Vk_E$ arranged in its columns.

\section{Results}
\subsection{Experimental setup}
Our objective is to analyze how the proposed method performs compared to state-of-the-art approaches in terms of classification accuracy, required training data, and robustness in out-of-distribution scenarios in limited training data setting. To achieve this, we created training sets of varying sizes from the original training set for each dataset under examination. We then trained the models using these training sets and assessed their performance on the original test set. Each train split was generated by randomly selecting (without replacement) samples from the original training set, and we repeated the experiments for each split size ten times. The same train-test data samples were used for all algorithms in each split.

In order to assess the effectiveness of the proposed approach, we utilized several comparison methods. These included PointNet \cite{qi2017pointnet}, DGCNN \cite{wang2019dynamic}, and multilayer perceptron (MLP) \cite{pedregosa2011scikit} in FSpool feature embedding space \cite{zhang2019fspool}. We also conducted a comparative analysis with various conventional machine learning techniques across different set feature embedding spaces. These included logistic regression (LR), kernel support vector machine (k-SVM), multilayer perceptron (MLP), and nearest subspace (NS) classifier models \cite{pedregosa2011scikit} in GeM1, GeM2, GeM4 \cite{radenovic2018fine}, COVpool \cite{wang2020deep,acharya2018covariance}, and FSpool \cite{zhang2019fspool} embedding spaces. The performance of the proposed method was evaluated in relation to these baselines. We conducted these evaluations in addition to performing out-of-distribution experiments. In the proposed method, we selected the number of basis vectors for the subspaces $\Vk_E$ such that the total variance explained by the chosen basis vectors in the $k$-th class captured up to 99\% of the total variance explained by that class.

To assess the relative performance of the methods, we evaluated them on several datasets, including Point cloud MNIST \cite{pcmnist,lecun1998gradient}, ModelNet \cite{wu20153d}, and ShapeNet \cite{chang2015shapenet} datasets. We additionally applied random translations, anisotropic scaling, and shear transformations to both the training and test sets of the datasets. For the ShapeNet dataset, we tested the methods under two experimental setups: the regular setup, where both the training and test sets contained point sets at the same deformation magnitude level, and the out-of-distribution setup, where the training and test sets contained point sets at different deformation magnitude levels.

\begin{figure*}[]
	\centering
	\includegraphics[width=0.85\textwidth]{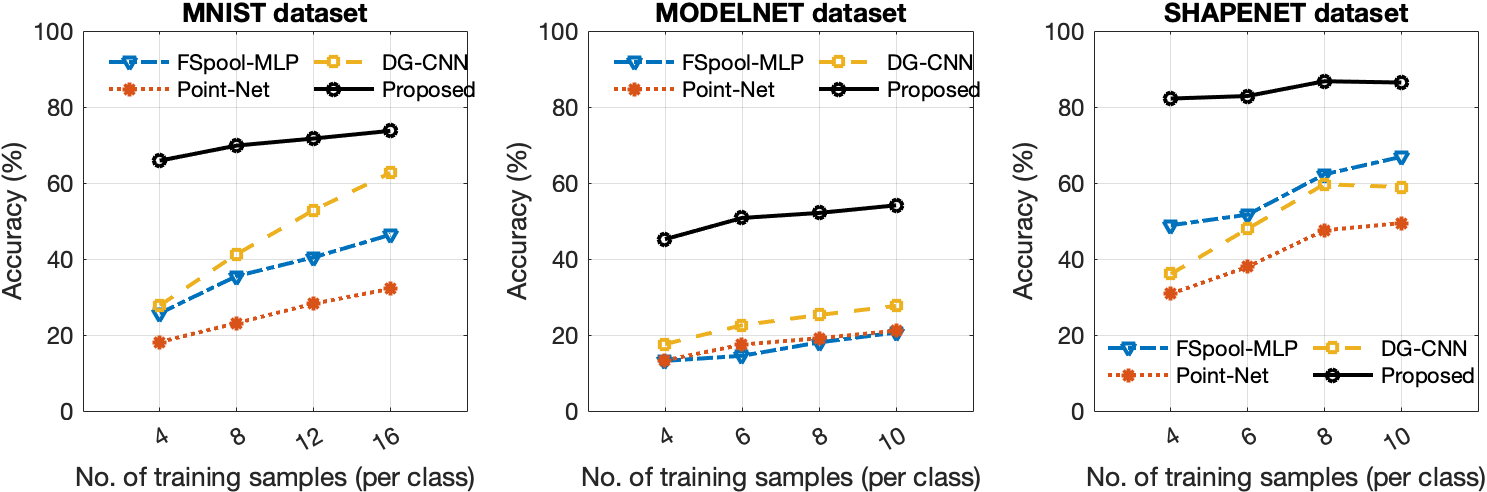}
	\caption{The accuracy of different methods as a function of the number of training samples per class, evaluated on MNIST, ModelNet, and ShapeNet datasets.}
	\label{resfig01_aim2}
\end{figure*}

\subsection{Accuracy in synthetic case}
We first evaluated the effectiveness of the proposed method by comparing it with other state-of-the-art techniques on two synthetic datasets. The synthetic datasets were generated by selecting one sample per class from the point cloud MNIST and ShapeNet datasets, followed by introducing random translations, anisotropic scaling, and shear transformations to each selected sample to generate training and test sets. Specifically, the training set consisted of two samples per class, while the test set comprised 25 samples per class. The obtained comparative results are displayed in Fig.~\ref{resfig00_aim2}. As observed, the proposed method substantially outperformed the other methods in this synthetic scenario.

\begin{figure}[]
	\centering
	\includegraphics[width=0.35\textwidth]{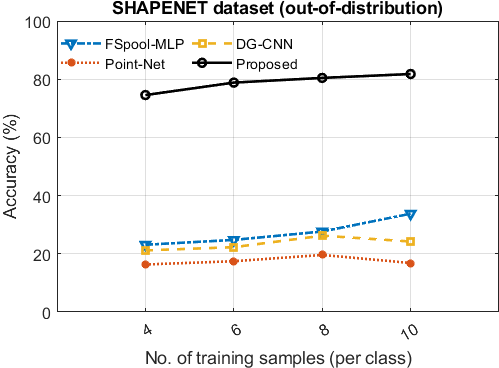}
	\caption{Performance assessment under an out-of-distribution experimental setup where training and test distributions vary in terms of deformation magnitudes. The performance of the methods was assessed in terms of percentage test accuracy and plotted against the number of training images per class.}
	\label{resfig02_aim2}
\end{figure}

\subsection{Accuracy and efficiency in real datasets}
We conducted the performance evaluation of the proposed method by comparing it with several state-of-the-art techniques, including PointNet, DGCNN, and MLP in FSpool feature embedding space, on the MNIST, ShapeNet, and ModelNet datasets. Fig.~\ref{resfig01_aim2} presents the average test accuracy values obtained for different numbers of training samples per class. The results demonstrate that our proposed method outperformed the other methods across the range of training sample sizes used to train the models. Notably, the proposed method's accuracy vs. training size curves exhibited a smoother trend in most cases compared to the other methods.

\subsection{Out-of-distribution robustness}
To assess the effectiveness of the proposed method under the out-of-distribution setting, we introduced a gap between the magnitudes of deformations in the training and test sets. Specifically, we used $\G_{out}$ as the deformation set for the `out-distribution' test set, while $\G_{in}$ was the deformation set for the `in-distribution' training set. We trained the models using the `in-distribution' data and tested using the `out-distribution' data. For our out-of-distribution experiment, we used the ShapeNet dataset with small deformations as the `in-distribution' training set and the ShapeNet dataset with larger deformations as the `out-distribution' test set (see Fig.~\ref{resfig02_aim2}). The results show that the proposed method outperformed the other methods by an even more significant margin under the challenging out-of-distribution setup, as shown in Fig.~\ref{resfig02_aim2}. Under this setup, the proposed method obtained accuracy figures closer to that in the standard experimental setup (i.e., ShapeNet in Fig.~\ref{resfig01_aim2}). On the other hand, the accuracy of the other methods declined significantly under the out-of-distribution setup compared to the standard experimental setup (see ShapeNet results in Figs.~\ref{resfig01_aim2} and \ref{resfig02_aim2}).


\begin{figure*}[!hbt]
	\centering
	\includegraphics[width=1.0\textwidth]{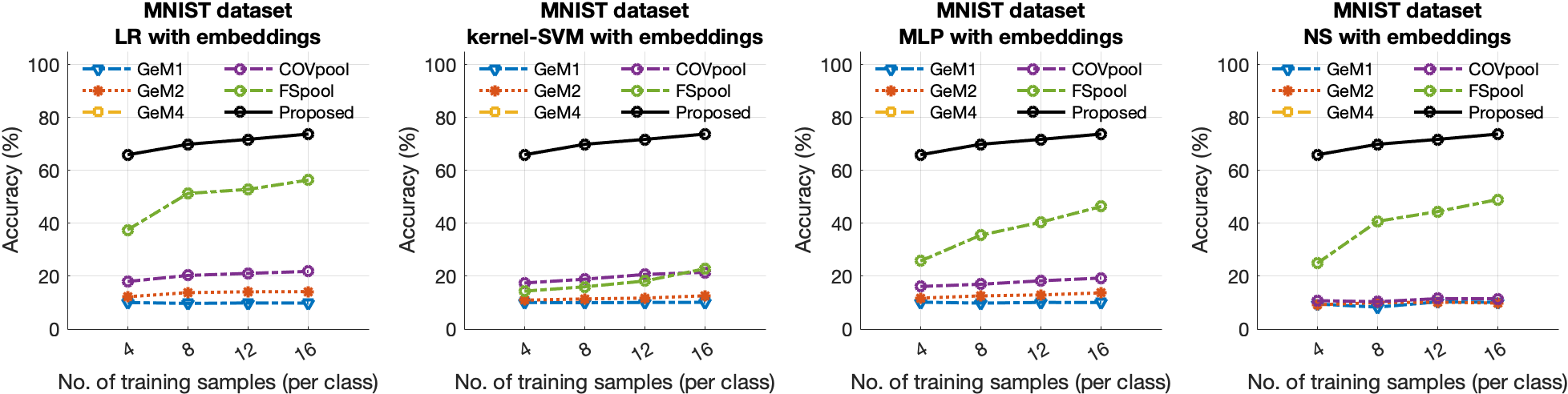}
	\caption{Comparative analysis of the percentage test accuracy results attained by the proposed method and the conventional machine learning techniques implemented across different feature embedding spaces.}
	\label{resfig04_aim2}
\end{figure*}




\subsection{Comparison with set-embedding-based methods}
We further evaluated the proposed method against various set embedding-based approaches in combination with classical machine learning methods. The study involved comparing the proposed method with different classifier techniques, including LR, k-SVM, MLP, and NS \cite{pedregosa2011scikit}, that were employed with various set-to-vector embedding methods, such as GeM (1,2,4) \cite{radenovic2018fine}, COVpool \cite{wang2020deep,acharya2018covariance}, and FSpool \cite{zhang2019fspool}. Fig.~\ref{resfig04_aim2} illustrates the percentage test accuracy results obtained from these modified experiments, along with the results of the proposed method for comparison. As shown in Fig.~\ref{resfig04_aim2}, the proposed method outperformed all these models in terms of test accuracy.


\section{Discussion}
This paper presents a new method for classifying point sets using linear optimal transport (LOT) subspaces. Our method is appropriate for problems where the data at hand can be represented as instances of prototype template point set patterns observed under smooth, nonlinear, and one-to-one transformations. 
The results achieved in different experimental scenarios indicate that our proposed approach can deliver accuracy results comparable to state-of-the-art methods, provided that the data adheres to the generative model specified in equation~\eqref{gen_model_aim2}. Additionally, the nearest LOT subspace technique was shown to be more data-efficient in these cases, meaning that it can attain higher accuracy levels using fewer training samples. 

Our proposed method maintains high classification accuracy, even in challenging out-of-distribution experimental conditions, as depicted in Fig.\ref{resfig02_aim2}, whereas the accuracy figures of other methods decline sharply. These results indicate that our method provides a better overall representation of the underlying data distribution, resulting in robust classification performance. The key to achieving better accuracy under out-of-distribution conditions is that our method not only learns the deformations present in the data but also learns the underlying data model, including the types of the deformations, such as translation, scaling, and shear, and their respective magnitudes. These deformation types can be learned from just a few training samples containing those deformations, as well as potentially from the mathematically prescribed invariances proposed in \cite{shifat2023invariance}.

Our proposed method, which utilizes the nearest subspace classifier in the LOT domain, is more suitable for classification problems in the above category compared with general set embedding methods in combination with classical machine learning classifiers, as demonstrated by its classification performance. Typically, point set data classes in their original domain do not constitute embeddings, and commonly used set-to-vector representation techniques are inadequate in generating effective embeddings for them, as indicated by the results. This presents a significant challenge for any machine learning approach to perform effectively. However, the subspace model is appropriate in the LOT domain since the LOT transform provides a linear embedding and convex data geometry. Moreover, considering the subspace model in the LOT space improves the generative nature of our proposed classification method by implicitly including the data points from the convex combination of the provided training data points. 

\section{Conclusions}
In this paper, we propose an end-to-end classification system designed for a specific category of point set classification problems, where a data class can be considered as a collection of instances of a template pattern observed under a set of spatial deformations. If these deformations are appropriately modeled as a collection of smooth, one-to-one, and nonlinear transformations, then the data classes become easily separable in the transform space, specifically the LOT space, due to the properties outlined in the paper. These properties also enable the approximation of data classes as convex subspaces in the LOT space, resulting in a more suitable data model for the nearest subspace method. As we observed in our experiments, this approach yields high accuracy and robustness against out-of-distribution conditions. Many point set classification problems can be formulated in this way, and therefore, our proposed solution has wide applicability.

Finally, we note that there can be many potential adaptations of the proposed method. For instance, the linear subspace method in the presented LOT space could be adjusted to incorporate alternative assumptions regarding the set that best represents each class. While some problems might benefit from a linear subspace method similar to the one described earlier, where all linear combinations are allowed, other problems may be require constraining the model using linear convex hulls. Additionally, investigating the sliced-Wasserstein distance using discrete CDT transform (as proposed in \cite{zhuang2022local}) in conjunction with subspace models is another promising avenue for future research.

Our proposed approach provides promising results in point set classification and serves as a basis for further exploration in this domain. As the amount of 3D (or N-D) data continues to increase and accurate object recognition and scene understanding become more crucial, we believe that the combination of linear optimal transport embeddings and subspace modeling in the transform space will become increasingly significant in this context. We anticipate that our proposed method will inspire further research in this direction and lead to novel developments in recognizing 3D (or N-D) objects or distributions.

\section*{Acknowledgments}
This work was supported in part by NIH grant GM130825, NSF grant 1759802, and CSBC grant U54-CA274499.

\bibliographystyle{ieeetr}
\bibliography{refs}


\end{document}